\documentclass[letterpaper, 10 pt, journal, twoside]{ieeetran}
\usepackage[caption=false,font=normalsize,labelfont=sf,textfont=sf]{subfig}
\usepackage[T1]{fontenc}
\usepackage{aecompl}
\usepackage{epstopdf}
\usepackage{cite}
\usepackage{amsmath,amssymb,amsfonts}
\usepackage{algorithmic}
\usepackage{algorithm}
\usepackage{graphicx}
\usepackage{textcomp}
\usepackage{xcolor}
\usepackage{float}
\usepackage{graphicx}
\usepackage{picinpar}
\usepackage{babel}
\usepackage{amsmath}
\usepackage{url}
\usepackage[latin1]{inputenc}
\usepackage{colortbl}
\usepackage{soul}
\usepackage{multirow}
\usepackage{pifont}
\usepackage{color}
\usepackage{alltt}
\usepackage[hidelinks]{hyperref}
\usepackage{enumerate}
\usepackage{siunitx}
\usepackage{breakurl}
\usepackage{epstopdf}
\usepackage{pbox}
\usepackage{authblk}
\usepackage{babel}
\usepackage{mathrsfs,balance}
\usepackage{tcolorbox}
\usepackage{fancyhdr}
\usepackage{lipsum}
\usepackage{threeparttable}
\usepackage{booktabs}
\usepackage{makecell}

\newcommand{\bobs}{{\mathbf{o}}}
\newcommand{\bact}{{\mathbf{a}}}
\newcommand{\beps}{{\boldsymbol{\epsilon}}}

\begin{document}


\title{VPWEM: Non-Markovian Visuomotor Policy \\ with Working and Episodic Memory
}

\author{Yuheng Lei, Zhixuan Liang, Hongyuan Zhang, Ping Luo
\thanks{Manuscript received March 5, 2026; revised June 24, 2026; accepted July 22, 2026. This paper was recommended for publication by Editor Wei Pan upon evaluation of the Associate Editor and Reviewers' comments.}
\thanks{All authors are affiliated to School of Computing and Data Science, University of Hong Kong. Correspondence should be sent to Yuheng Lei and Ping Luo with email: \tt\footnotesize leiyh@connect.hku.hk;pluo@hku.hk}
\thanks{Digital Object Identifier (DOI): see top of this page.}

}

\markboth{IEEE ROBOTICS AND AUTOMATION LETTERS, Accepted July 2026}%
{Lei \MakeLowercase{\textit{et al.}}: VPWEM: Non-Markovian Visuomotor Policy with Working and Episodic Memory}



\maketitle

\begin{abstract}

Imitation learning from human demonstrations has achieved significant success in robotic control, yet most visuomotor policies still condition on single-step observations or short-context histories, making them struggle with non-Markovian tasks that require long-term memory. Simply enlarging the context window incurs substantial computational and memory costs and encourages overfitting to spurious correlations, leading to catastrophic failures under distribution shift and violating real-time constraints in robotic systems. By contrast, humans can compress important past experiences into long-term memories and exploit them to solve tasks throughout their lifetime. In this paper, we propose VPWEM, a non-Markovian visuomotor policy equipped with working and episodic memories. VPWEM retains a sliding window of recent observation embeddings as short-term working memory, and introduces a Transformer-based contextual memory compressor that recursively converts out-of-window observations into a fixed number of episodic memory embeddings. The compressor uses self-attention over a cache of past summary embeddings and cross-attention over a cache of historical observations, and is trained jointly with the policy. We instantiate VPWEM on diffusion policies to exploit both short-term and episode-wide information for action generation with nearly constant memory and computation per step. Experiments demonstrate that VPWEM outperforms state-of-the-art baselines including diffusion policies and vision-language-action (VLA) models by more than 20\% on the memory-intensive manipulation tasks in MIKASA and achieves an average 5\% improvement on the mobile manipulation benchmark MoMaRT. Code is available at \url{https://github.com/HarryLui98/code_vpwem}.
\end{abstract}


\section{Introduction}
\label{sec:intro}
\begin{figure}[t]
    \centering
    \includegraphics[width=0.8\textwidth,keepaspectratio=true,trim=158 380 300 130,clip]{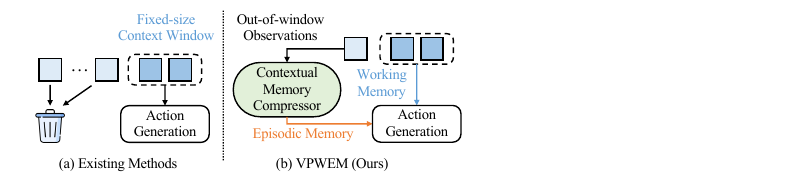}
    \vspace{-0.25in}
    \caption{Comparison between VPWEM and existing methods: (a) Policies typically predict actions based on observations within a fixed-size context window. Historical observations that move out of the window are typically discarded and will not be used during action generation. (b) VPWEM augments diffusion-based policies with working and episodic memories, where the contextual memory compressor recursively consolidates historical observations into fixed-size memory embeddings.}
    \vspace{-0.25in}
\label{fig:framework-intro}
\end{figure}

\IEEEPARstart{I}{mitation} learning from demonstrations has achieved significant success in learning high-performing robot policies \cite{ravichandar2020recent, kroemer2021review}. While model architectures and training datasets have been scaled up to the level of several millions or even billions, existing policies typically predict actions based on single-step observations \cite{nair2023r3m, radosavovic2023real, zitkovich2023rt, reuss2024multimodal, kim2024openvla, black2025pi0} or utilize observation histories with short context lengths, such as 2 \cite{ghosh2024octo, chi2024diffusion, ze20243d, liu2025rdt}, 6 \cite{brohan2023rt}, and 10 \cite{huang2025otter}, as shown in Figure \ref{fig:framework-intro}(a). However, robotic tasks in realistic scenarios are often non-Markovian due to sensor limitations, environmental stochasticity, and long-horizon structures involving multiple subgoals. Similarly, humans make decisions not only based on the current observation but also on past experiences in mind. Policies without memory may struggle to capture such long-term temporal dependencies, leading to execution failures.

Despite the conceptual importance of long-term memories for robot policies, existing approaches limit history length in practice due to two key challenges: (1) Computational complexity and memory requirement grows rapidly as history length $L$ increases, \textit{e.g.}, the $O(L^2)$ complexity of self-attention mechanism \cite{vaswani2017attention}, which makes training prohibitively expensive and leads to significant inference latency. Although some recent work \cite{jia2025x} has preliminarily explored the potential of model with $O(L)$ complexity \cite{katharopoulos2020transformers, gu2023mamba, dao2024transformers} in robot learning, they mainly focus on increasing efficiency within short context length (e.g., 5 in MaIL \cite{jia2024mail}) and thus still lack the capability for long-term memory. 
(2) Naively conditioning on a longer history without careful consideration often lead to performance degradation rather than improvement. \cite{de2019causal} frame this issue as a causal confusion problem arising from nuisance correlations. \cite{wen2020fighting} further highlight the copycat problem, where the agent tends to replicate the previous action. When the agent develops an incorrect dependence on nuisance variables in the history, it may achieve low training error but suffer from catastrophic failures under distributional shifts. Various solutions have been proposed to resolve this issue, including adversarial training \cite{wen2020fighting}, utilizing auxiliary priming information \cite{wen2022fighting}, and adding past action leakage regularization \cite{seo2023regularized}. 

Hence, naively conditioning on an ever-growing observation history during task execution is impractical. Instead, neuroscience reveals that the hippocampus in human brain continually converts working memories into long-term storage in the cortex, enabling lifelong retention of knowledge and experience despite the brain's limited volume \cite{mcclelland1995there}. Inspired by this mechanism, similar ideas have been adopted in long document understanding \cite{chevalier2023adapting, fang2025artificial, yu2025memagent} and long video understanding \cite{he2024ma, balazevic2024memory}. 
In this paper, we propose VPWEM, which augments diffusion policies with working and episodic memory. As illustrated in Figure \ref{fig:framework-intro}(b), the key idea is to employ a contextual memory compressor that recursively consolidates out-of-window observations into fixed-size episodic memory embeddings, analogous to the role of the hippocampus in the human brain. This allows the policy to maintain episodic memory with bounded computational and memory costs. Through end-to-end optimization with the behavior cloning objective, the compressor is encouraged to preserve task-relevant information while reducing sensitivity to nuisance correlations.
We conduct extensive experiments across three benchmarks. VPWEM outperforms state-of-the-art baselines by more than 20\% on memory-intensive manipulation tasks in MIKASA \cite{cherepanov2025memory}, achieves an average 5\% gain on the mobile manipulation benchmark MoMaRT \cite{wong2022error}, and performs on par with baselines on the (almost) Markovian Robomimic benchmark \cite{mandlekar2022matters}. A real-robot experiment further demonstrates the practical effectiveness of VPWEM.

In summary, our main contributions are three-fold: (1) We propose a novel framework that employs a memory compressor to recursively condense historical observations into fixed-size memory embeddings, which serve as a dynamic summary of the trajectory; (2) We instantiate this idea on two diffusion policy baselines, DP \cite{chi2024diffusion} and MaIL \cite{jia2024mail}, redesigning their training and inference pipelines so that action generation is conditioned on both long- and short-term memory; (3) We conduct extensive experiments showing that our framework substantially improves performance in memory-intensive tasks, while achieving performance that is on par with baselines in Markovian tasks.

\section{Related Work}
\label{sec:related}

\textbf{Long-context Models.} 
Effectively handling long contexts has been a long-standing challenge, motivating a range of techniques to address this issue \cite{huang2023advancing, liu2025comprehensive}. Extrapolation methods, such as RoPE \cite{su2024roformer}, modify positional embeddings to extend the context window of pre-trained models. Sparse attention approaches \cite{beltagy2020longformer} reduce the quadratic cost of vanilla attention by masking irrelevant attention computations. Memory-augmentation methods introduce explicit external memory with caching, compression, and retrieval: Memorizing Transformer \cite{wumemorizing} employ $k$-nearest neighbor to retrieve semantically related content from a key-value database, while RMT \cite{bulatov2022recurrent} integrates learnable memory tokens to facilitate information transfer across segments. Other lines of work adopt architectures with linear complexity, such as recurrent neural networks (RNNs) \cite{beck2024xlstm} and state space models (SSMs) \cite{gu2023mamba}. Beyond language modeling, similar techniques have shown considerable promise in long-context video understanding and generation \cite{he2024ma, balazevic2024memory, zhang2025packing}. Despite these advancements, the integration of long-term contextual memories into robot policies remains largely unexplored, and is the primary focus of this work.

\textbf{Imitation Learning.}
Imitation learning from human demonstrations is a simple yet effective approach for robot learning \cite{ravichandar2020recent, kroemer2021review}. Early methods often used MLP policies under a Markov Decision Process (MDP) formulation, while recent works stack limited past frames and use sequence models such as Transformers \cite{vaswani2017attention} to handle partial observability. This Markovian or short-history setting has enabled many vision-language-action models (VLAs), including RT-1/2 \cite{brohan2023rt, zitkovich2023rt}, Otter \cite{huang2025otter}, OpenVLA \cite{kim2024openvla}, $\pi_0$ \cite{black2025pi0}, and RDT \cite{liu2025rdt}, but can struggle with long-range temporal dependencies.
Several research directions have explored how to incorporate long-term memory into imitation learning.
One line maintains belief-like internal states using RNNs \cite{mandlekar2022matters} or SSMs \cite{zhou2025mtil}, with MGDP \cite{huang2025memory} further leveraging Mamba architecture with gated attention to capture long-term dependencies. However, these methods propagate history through step-by-step recurrent state updates, whose fixed-dimensional internal states can limit long-term information retention. In contrast, VPWEM represents long-term memory as a small set of episodic memory embeddings and updates them through attention over both the summary and observation caches, enabling selective consolidation of retained historical observations and past memory embeddings. Another line directly uses the full history of interleaved observation and action tokens for autoregressive action prediction, which captures extensive context but becomes slow as sequence length grows and depends heavily on action tokenizer design \cite{fu2024context, vuong2025action}. DP-PTP \cite{torne2025learning} also shows that diffusion policies can benefit from past-action prediction as an auxiliary loss.
Recent works on VLAs also address long-term memory from different perspectives: CronusVLA \cite{li2025cronusvla} extends single-frame VLA to multi-frame prediction via post-training; ContextVLA \cite{jang2025contextvla} compresses multi-frame VLM features by average pooling; MemoryVLA \cite{shi2025memoryvla} retrieves and fuses decision-relevant memory entries for action generation; BPP \cite{mark2026bpp} uses an off-the-shelf VLM to select task-relevant historical keyframes; MemER \cite{sridhar2025memer} fine-tunes a high-level VLM to select keyframes and generate textual subtask goals; and MEM \cite{torne2026mem} combines video-based short-horizon memory and text-based long-horizon memory. Our method instead explicitly decouples long-term memory modeling from single-frame representation learning. The single-frame encoder extracts rich features from the current observation, while the external memory compressor maintains a compact episode-level memory of out-of-window observations. This modular design naturally supports integration with various visuomotor policies, including VLAs, by providing compressed episodic memory embeddings as additional context.

\textbf{Diffusion Policies.} Diffusion policies have recently gained significant attention for their impressive ability to model multi-modal distributions in high-dimensional action spaces \cite{chi2024diffusion}. A diffusion policy generally contains two parts \cite{dong2024cleandiffuser}: (i) the condition backbone that encodes multi-modal perception, \textit{e.g.}, proprioceptive states, language instructions \cite{zhang2024language}, 2D RGB images \cite{chi2024diffusion}, and 3D point clouds \cite{ze20243d}, into low-dimensional condition embeddings and (ii) the diffusion backbone that iteratively predict the noise from noisy actions given conditional embeddings, with optional architecture including MLP \cite{pearceimitating}, U-Net \cite{chi2024diffusion, ze20243d}, and Transformer \cite{liu2025rdt, reuss2024multimodal}. Prior work mainly focuses on strengthening the condition backbone by using more advanced pretrained vision-language models (VLMs) to extract more informative latent features, or on improving the efficiency of the diffusion backbone for action generation via rectified flow \cite{zhang2025flowpolicy} or discrete diffusion \cite{liang2025discrete}. In contrast, our method is orthogonal to these efforts: we explicitly insert a memory module to enhance temporal understanding in non-Markovian tasks.

\begin{figure*}[t]
    \centering
    \includegraphics[width=0.95\textwidth,keepaspectratio=true,trim=140 280 115 50,clip]{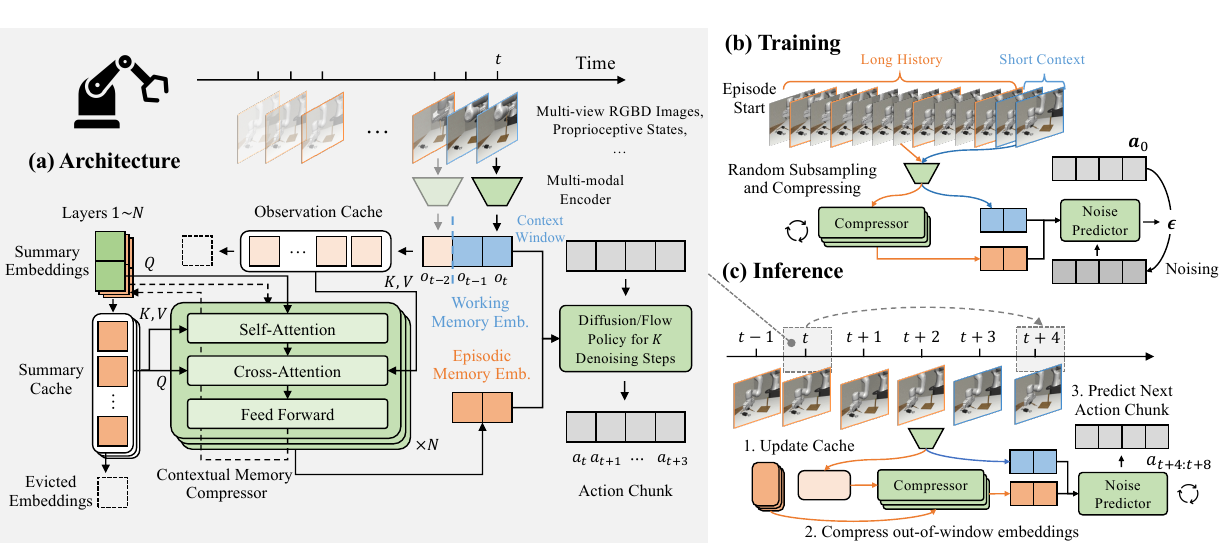}
    \vspace{-0.15in}
    \caption{Overview of the VPWEM framework. (a) Policy architecture. Modules in green are learnable components, including the multi-modal encoder, contextual memory compressor, summary embeddings, and the Transformer-based noise predictor in the diffusion policy. These components are optimized end-to-end with a behavior cloning loss. (b) Training process. Each training sample contains the complete trajectory from the beginning of the episode. The short-term memory component follows the standard diffusion policy baselines. Observations outside the context window are subsampled with a fixed ratio and passed to the contextual memory compressor, which outputs a fixed number of memory embeddings. The combined long- and short-term memory embeddings are used to condition the noise prediction network. (c) Inference process. Each decision step consists of three steps: encoding new frames and updating the observation and summary caches; compressing out-of-window embeddings to obtain long-term memories; and predicting the action chunk.}
    \vspace{-0.2in}
\label{fig:framework}
\end{figure*}

\section{Problem Formulation}
\label{sec:prel.formulation}
Following standard imitation learning settings, we formulate a robotic task as a Partially Observable Markov Decision Process (POMDP) $\mathcal{M}=(\mathcal{S}, \mathcal{A},  \mathcal{P}, \mathcal{R}, \Omega, \mathcal{O})$, where $\mathcal{S}$ and $\mathcal{A}$ is the state space and action space for tasks, $\Omega$ is the observation space of the robots, such as RGB images and proprioceptive states, $\mathcal{P}:\mathcal{S}\times\mathcal{A}\rightarrow\mathcal{S}$ is the transition dynamics, $\mathcal{O}:\Omega\rightarrow\mathcal{S}$ is the observation emission function, and $\mathcal{R}$ is the reward function of the task. The optimal policy $\pi_\theta(a_t|h_t)$ necessitates conditioning on the full history $h_t=o_{\leq t}=(o_0,\cdots,o_t)$ to predict actions, while in practice existing methods often limit the context window to $L$ and use the policy with form of $\pi_\theta(a_t|o_{t-L:t})$. We consider to learn the robot policy $\pi_\theta$ from an expert demonstration dataset $\mathcal{D}=\{\tau_n\}_{n=1}^{N}$, where each expert trajectory $\tau$ contains a sequence of observation-action pairs $(o_0,a_0,\cdots,o_T)$. The goal is to mimic expert behaviors by optimizing policy parameters $\theta$.

\section{Method}
\label{sec:method}

The overview of VPWEM is illustrated in Figure \ref{fig:framework}. In addition to using observations within the context window as working meory (Section \ref{sec:dplscm.short}), we leverage a contextual memory compressor to distill essential information from observations that have fallen outside this window into fixed-size summary embeddings (Section \ref{sec:dplscm.long}). The resulting short-term working memory and long-term episodic memory provide complementary conditioning signals that jointly guide the action generation process (Section \ref{sec:dplscm.actiongen}).

\subsection{Working Memory}
\label{sec:dplscm.short}

Following the problem formulation in Section \ref{sec:prel.formulation}, we employ a multi-modal encoder to extract features from the various types of observations. Specifically, given $N_c$ cameras, we use $N_c$ image encoders to process the raw RGBD observations $I_{n_c,t} \in \mathbb{R}^{H \times W \times C}$, for $n_c \leq N_c$, producing visual features $o_{I,t} \in \mathbb{R}^{D_I}$. In parallel, we encode the low-dimensional proprioceptive states into $o_{P,t} \in \mathbb{R}^{D_P}$. These features are then concatenated and passed through an MLP to obtain a joint feature vector $o_t \in \mathbb{R}^D$ that captures the essential information of a single frame at time step $t$.

We define the working memory as the observation embeddings within a predefined sliding window, $w_{t}\triangleq o_{t-L:t}=\texttt{concat}(o_{t-L+1},\cdots,o_{t-1},o_t)\in\mathbb{R}^{L\times D}$, which is maintained using a first-in-first-out (FIFO) scheme. This design alleviates the quadratic computational cost and mitigates overfitting, and is therefore widely adopted in prior work. However, discarding earlier historical information makes it insufficient for non-Markovian tasks that require dependencies extending beyond the fixed context window $L$, thereby motivating the need for complementary long-term episodic information.

\subsection{Episodic Memory}
\label{sec:dplscm.long}

While long-term information resides in the observation history $o_{\leq t-L}=\texttt{concat}(o_0,o_1,\cdots,o_{t-L})$, naively conditioning on this ever-growing observation history is impractical. Hence, we employ a Transformer-based compressor that recursively consolidates out-of-window observations into long-term contextual memory embeddings. 

Formally, given a single-frame observation embedding $o_\tau$ that has just exited the short-term context window, we further incorporate temporal information by adding a positional embedding (PE) $f_\tau=o_\tau+\texttt{PE}(\tau)\in\mathbb{R}^D$, where $\tau$ denotes the timestep of the frame within the episode. The embedding is then pushed into the out-of-window observation cache $\mathcal{C}_f\leftarrow\mathcal{M}(\mathcal{C}_f\cup f_\tau)$ for future retrieval, where $\mathcal{M}$ is a cache management method that maintains a maximum cache size $S$, such as FIFO. 

To extract essential information from long observation histories, we employ a contextual memory compressor that follows the standard Transformer encoder architecture \cite{vaswani2017attention}. Each block within this compressor comprises a self-attention layer for interactions with past summary embeddings, a cross-attention layer for interactions with observation embeddings, and a feed-forward network. Layer normalization and residual connections are applied before or after each of these sub-modules. The contextual memory compression process can be formally presented as:
\begin{align} \label{eq:compress.query}
q_{1,\tau} &= q, \\ \label{eq:compress.selfattn}
x_{1,n} &= q_{n,\tau} + \texttt{attn}(q_{n,\tau}Q_{s}, \bar{\mathcal{C}}_{q,n}K_{s},\bar{\mathcal{C}}_{q,n}V_{s}), \\ \label{eq:compress.crossattn}
x_{2,n} &= x_{1,n} + \texttt{attn}(x_1Q_{c}, \bar{\mathcal{C}}_{f}K_{c},\bar{\mathcal{C}}_{f}V_{c}),\!\! \\
q_{n+1,\tau} &= x_{2,n} + \texttt{MLP}(x_{2,n}), \\
e_{\tau} &= \texttt{MLP}(q_{N,\tau}).
\end{align}
Specifically, each transformer block indexed by $n\!\leq\!N$ operates on $M$ queries, $q_{n,\tau}\!\in\!\mathbb{R}^{M\times D}$. The initial queries for the first layer $q_{1,\tau}$ are trainable model parameters $q$. Similar to the observation cache, a summary cache $\mathcal{C}{q,n}$ is maintained per block, storing query embeddings from previous timesteps and updated at each step as $\mathcal{C}_{q,n}\!\leftarrow\!\mathcal{M}(\mathcal{C}_{q,n}\cup q_{n,\tau})$. We concatenate the observation cache into $\bar{\mathcal{C}}_{f}=\texttt{concat}(\mathcal{C}_{f})\in\mathbb{R}^{S\times D}$ and the summary cache into $\bar{\mathcal{C}}_{q}=\texttt{concat}(\mathcal{C}_{q})\in\mathbb{R}^{S\times M \times D}$. Note that $S$ denotes the cache size shared by the observation and summary caches, whereas $M$ denotes the number of episodic memory embeddings. The self-attention mechanism in Eq. (\ref{eq:compress.selfattn}) uses $q_{n,\tau}$ to query $\bar{\mathcal{C}}_{q}$, producing an output $x_{1,n}$ that compactly summarizes past memories up to the current timestep. $x_{1,n}$ then serves as queries for a cross-attention mechanism in Eq. (\ref{eq:compress.crossattn}), which attends to multi-frame observation features from $\bar{\mathcal{C}}_{f}$, thereby capturing long-term dependencies across frames. The block's output then becomes the input queries for the subsequent block. Finally, the output of the final layer, $q_{N,\tau}$, is projected to yield the episodic memory $e_{\tau}\in\mathbb{R}^{M\times D}$.

\subsection{Action Generation with Working and Episodic Memory}
\label{sec:dplscm.actiongen}

The core idea behind diffusion models \cite{ho2020denoising} is to iteratively transform a simple noise distribution into a complex target distribution through a sequence of denoising steps, optionally conditioned on contextual information. When applied tn robotic tasks \cite{chi2024diffusion}, the conditioning context is typically the robot's observation history \cite{chi2024diffusion, torne2025learning}. In VPWEM, at each timestep $t$, action generation is conditioned by both working memory $w_{t}$ (Section \ref{sec:dplscm.short}) and episodic memory $e_{\tau}$ (Section \ref{sec:dplscm.long}).
Starting from an initial Gaussian noise sample $\bact_K\sim\mathcal{N}(\mathbf{0},\mathbf{I})$, the diffusion backbone progressively denoises the sequences over $K$ timesteps, ultimately producing the final output $\bact_0$, which is used as the robot's action chunk $\bact_{t:t+H}$. Note that $L$ is the observation horizon and $H$ is the action prediction horizon. We illustrate the training and inference process of VPWEM in Figure \ref{fig:framework}. 

\textbf{Training.} The training of diffusion policies involves two main processes: the forward (noising) process and the reverse (denoising) process. In the forward process, we randomly sample examples $(\bobs_{\leq t},\bact_{t:t+H})$, abbreviated as $(\bobs,\bact_0)$, from the demonstration dataset. 
Unlike methods that use a fixed context window, our framework leverages the full trajectory history for action generation, resulting in observation inputs of varying lengths.
To handle this, we first sort samples by their positions within an episode and group them by similar sequence lengths, so that samples within a mini-batch have comparable input sizes. Although slight mismatches still remain, we pad the beginning of each episode with its first frame to enable parallel batch training of the policy. To increase robustness, we divide each trajectory into segments with a specified subsample ratio and randomly select one frame from each segment to construct the input to the contextual memory compressor. In addition, we detach $f_\tau$ and $q_{n,\tau}$ from the computational graph before storing them in the observation and summary caches. This prevents gradients from backpropagating through time, ensuring that historical information is propagated solely via the summary embeddings, substantially reducing memory consumption and mitigating overfitting issue. 

For each sampled $(\bobs,\bact_0)$, we can generate noisy action sequences $\bact_k=\sqrt{\bar\alpha_k} \mathbf{a}_0 + \sqrt{1-\bar\alpha_k}\beps$ at any denoising timestep $k\in[1,K]$, where $\bar{\alpha}_k$ and $\alpha_k$ are functions of $k$ that depend on the noise scheduler and $\beps\sim\mathcal{N}(\mathbf{0},\mathbf{I})$. In the reverse process, a neural network $\beps_\theta$ is trained to predict the noise $\beps$ from the noisy input $\bact_k$, conditioned on the working memory $w_{t}$, episodic memory $e_{\tau}$, and timestep $k$. The network employs a transformer decoder architecture and applies a cross-attention mask to condition the action chunk generation \cite{chi2024diffusion, dong2024cleandiffuser}. The training objective is:
\begin{equation}
    \!\mathcal{L} \!=\! \mathbb{E}_{(\bobs,\bact_0), \beps} \left\Vert\beps \!-\! \beps_\theta(\sqrt{\bar\alpha_k} \mathbf{a}_0 \!+\! \sqrt{1-\bar\alpha_k}\beps, w_{t}, e_{\tau}, k)\right\Vert^2_2. \label{eq:prel.trainingloss}
\end{equation}

Note that $e_{\tau}$ is the output of the contextual memory compressor, as defined in Eqs. (1)-(5). Consequently, the compressor is optimized jointly with the behavior cloning loss to extract task-related information from the history.

\textbf{Inference.} During inference, we maintain a queue of observation embeddings that provides contextual information from previously observed data, enabling the multi-modal encoder to focus exclusively on unseen frames. Observations within the context window serve as working memory, while those that move beyond the window are compressed into episodic memory by the well-trained contextual memory compressor. The resulting long-term episodic memory and short-term working memory are then used to predict the action chunk. The denoising network can iteratively transforms the noise sequence $\bact_K$ into $\bact_{K-1},\bact_{K-2},\cdots,\bact_{2},\bact_1$, and finally the robot action $\bact_0$, through $K$ steps of denoising:
\begin{equation}
    \bact_{k-1} \!=\! \frac{1}{\sqrt{\alpha_k}}\left( \bact_k \!-\! \frac{1-\alpha_k}{\sqrt{1-\bar{\alpha}_k}} \beps_\theta(\bact_k, w_{t}, e_{\tau}, k) \right) + \sigma_k \mathbf{z}. \label{eq:prel.dpinference}
\end{equation}
where $\mathbf{z}\sim\mathcal{N}(\mathbf{0},\mathbf{I})$, and $\sigma_k$ is also a function of $k$ derived from the noise scheduler.

\section{Experiments}
\label{sec:exp}

\subsection{Experimental Setup}

\begin{figure}[htb]
    \centering
    \vspace{-0.1in}
    \includegraphics[width=0.5\textwidth,keepaspectratio=true,trim=90 240 75 160,clip]{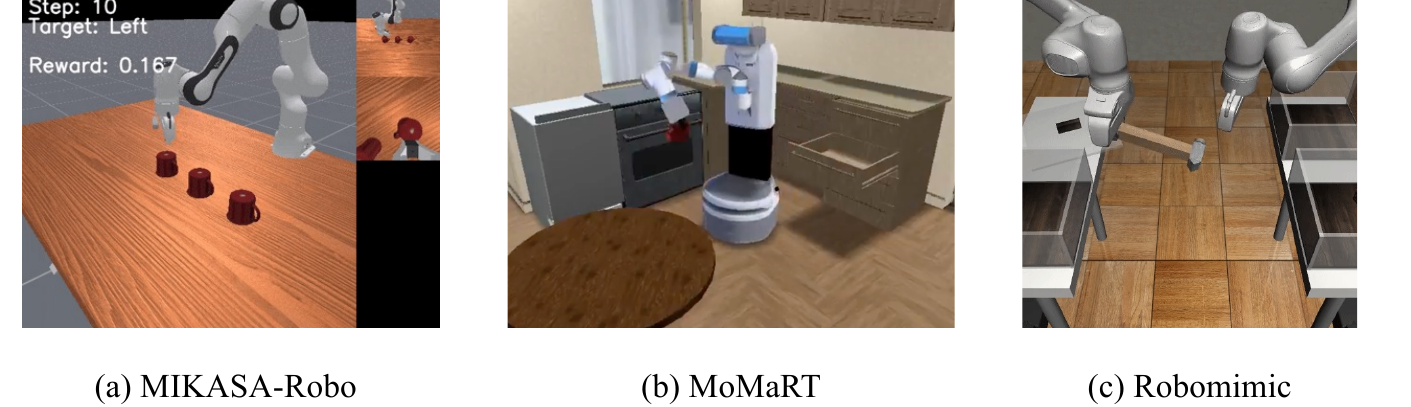}
    \vspace{-0.25in}
    \caption{Three benchmarks in our experiments.}
    \vspace{-0.1in}
\label{fig:benchmark}
\end{figure}

\textbf{Benchmark.} We evaluate our method across three benchmarks, as shown in Figure \ref{fig:benchmark}:
\begin{itemize}
    \item MIKASA (Memory-Intensive Skills Assessment Suite for Agents) \cite{cherepanov2025memory}. We use two memory-intensive tabletop manipulation tasks,  \textit{ShellGameTouch-v0} and \textit{RememberColor3-v0}. The former requires the policy to memorize the position of the ball after some time being covered by the cups and then interact with the cup the ball is under, whereas the latter requires it to remember the color of the cube and select it among other colors.
    \item MoMaRT (Mobile Manipulation RoboTurk) \cite{wong2022error}. We use five long-horizon mobile manipulation tasks in a mobile simulated kitchen, which are inherently non-Markovian: \textit{Table Setup from Dishwasher / Dresser} (dwsetup/dssetup), \textit{Table Cleanup to Dishwasher / Sink} (dwclean/skclean), \textit{Unload Dishwasher to Dresser} (unload), with expert or suboptimal demonstrations.
    \item Robomimic \cite{mandlekar2022matters}. We use two tabletop manipulation tasks, \textit{Square} and \textit{Transport}, with proficient-human (ph) or multi-human demonstration (mh) demonstrations, which can be roughly regarded as Markovian. We also include a non-Markovian \textit{Long-Horizon Square} task from \cite{torne2025learning}, where the robot must place a block onto the farthest peg from its initial location. The noisy scripted demonstrations prevent the policy from inferring the goal using current pose information alone, thus requiring memory of the initial state. 
\end{itemize}

\textbf{Baselines.} We compare our method with the following representative imitation learning baselines:
\begin{itemize}
    \item Recurrent Neural Network Policy (RNN) \cite{mandlekar2022matters} uses a long short-term memory (LSTM) network as backbone, where the final layer hidden states are passed to a Gaussian Mixture Model (GMM) to predict action.
    \item Diffusion Policy (DP) \cite{chi2024diffusion} employs a Transformer-based denoising network conditioned only on observations within a fixed-size context window.
    \item Diffusion Policy with Past Token Prediction (DP-PTP) \cite{torne2025learning} shares the same architecture as DP but introduces Past-Token Prediction (PTP) as an auxiliary objective alongside predicting future action tokens.
    \item Mamba Imitation Learning (MaIL) with Diffusion Policy \cite{jia2024mail} improves over DP by using a Mamba-based denoising network, whose recurrent nature helps better extract important information from the history.
\end{itemize}

We also report the performance of Octo \cite{ghosh2024octo}, OpenVLA \cite{kim2024openvla}, $\pi_0$ \cite{black2025pi0}, SpatialVLA \cite{qu2025spatialvla}, CronusVLA \cite{li2025cronusvla} and MemoryVLA \cite{shi2025memoryvla} in the MIKASA benchmark using the results from \cite{cherepanov2025memory, shi2025memoryvla}. We further compare our method with BPP \cite{mark2026bpp} and MemER \cite{sridhar2025memer}, which use a high-level Qwen3-VL 2B model \cite{bai2025qwen3} to generate image-based and language-based summarization, respectively.

\textbf{Implementation Details.} We implement the baselines and our proposed method using the open-source CleanDiffuser framework \cite{dong2024cleandiffuser}. We follow the multi-stage training in \cite{torne2025learning} to use the frozen vision encoder from short-context policies when training long-context policies. We run all simulated experiments on a single NVIDIA RTX 4090 GPU. To ensure reproducibility and statistical significance, we use three random seeds (100, 200, 300) and report the mean and standard deviation of final performance across these runs. We follow the hyperparameter settings recommended for baselines \cite{dong2024cleandiffuser, jia2024mail, chi2024diffusion}, and perform ablation studies on the hyperparameters of our method, as described in Section \ref{sec:exp.ablation}. Hyperparameter details are summarized in Table \ref{tab:hyper}.

\begin{table}[b]
\setlength{\tabcolsep}{1pt}
\vspace{-0.15in}
    \caption{Hyperparameters of DP-VPWEM}
    \label{tab:hyper}
    \vspace{-0.15in}
    \begin{center}
    \begin{tabular}{lc}
    \hline
    \textbf{Hyperparameter}&\textbf{Value}\\
    \hline
    \emph{Diffusion Policy} & \\
    Diffusion Model & DDPM \cite{ho2020denoising} \\
    Sampling steps & 50 \\
    Action chunk length $H$ & 8 \\
    Temperature & 1.0 \\
    Number of layers & 8 \\
    Embedding dimension $D$ & 256 \\
    \hline
    \emph{Contextual Memory Compressor}\\
    Number of layers $N$ & 2 \\
    Number of short-term embeddings $L$ & 2 \\
    Number of long-term embeddings $M$ & 2 \\
    Maximum cache size & 8 \\
    Dropout ratio of memory embeddings & 0.3 \\
    Subsample ratio &  5 (MIKASA) / 20 (otherwise) \\
    \hline
    \emph{Training}\\
    Batch size & 64 \\
    Learning rate & 1e-4 \\
    EMA rate & 0.999 \\
    Gradient steps & 6e5 (MIKASA) / 1e6 (otherwise) \\
    Optimizer & AdamW($\beta_1$ = 0.9, $\beta_2$ = 0.999) \\
    \hline
    \end{tabular}
    \end{center}
    \vspace{-0.05in}
\end{table}

\begin{figure*}[t]
    \centering
    \includegraphics[width=\textwidth,keepaspectratio=true,trim=20 400 20 20,clip]{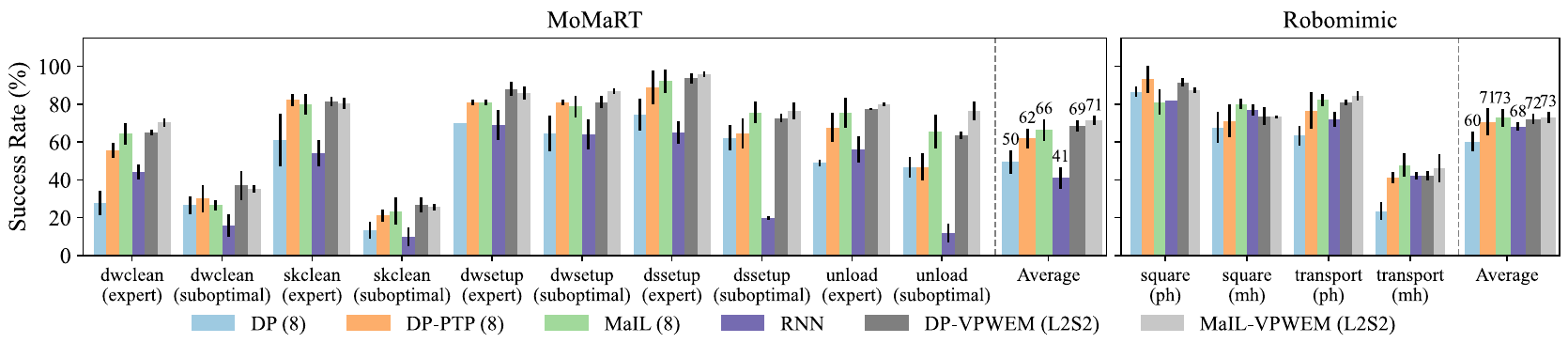}
    \vspace{-0.25in}
    \caption{Performance on MoMaRT and Robomimic benchmark.}
    \vspace{-0.1in}
\label{fig:momart}
\end{figure*}

\begin{figure}[b]
    \centering
    \vspace{-0.15in}
    \includegraphics[width=0.49\textwidth,keepaspectratio=true,trim=25 570 20 20,clip]{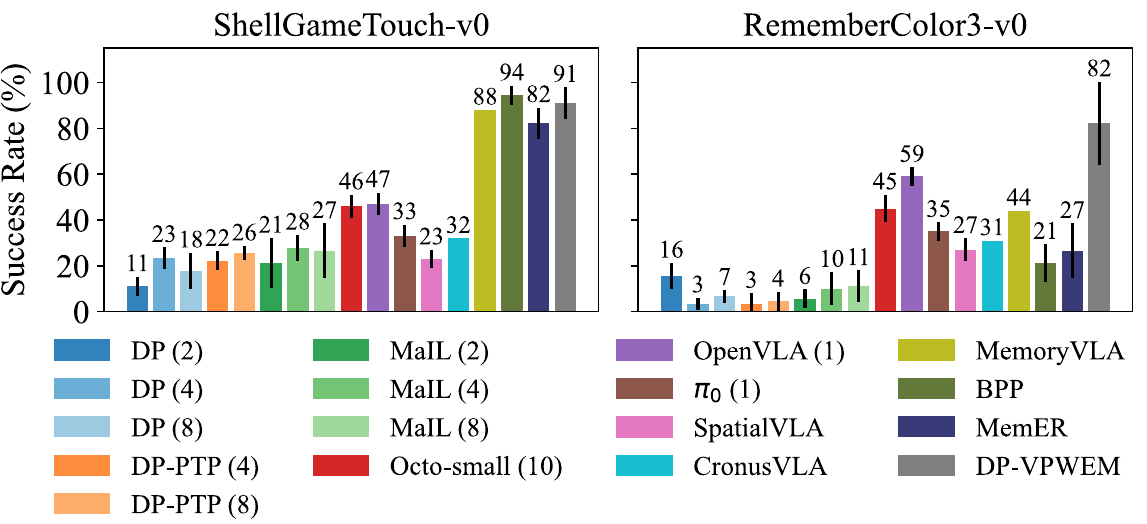}
    \vspace{-0.25in}
    \caption{Performance on MIKASA benchmark.}
\label{fig:mikasa}
\end{figure}

\begin{table}[h]
\setlength{\tabcolsep}{1pt}
    \caption{Comparison of Computational and Storage Costs on the unload-suboptimal task in the MoMaRT benchmark}
    \label{tab:computation}
    \vspace{-0.1in}
    \begin{center}
    \begin{tabular}{cccccccc}
    \hline
    \textbf{Methods} & \multicolumn{6}{c}{\textbf{DP-PTP}} & \textbf{DP-VPWEM} \\
    Context Length & 4 & 8 & 16 & 32 & 64 & 128 & L2S2 \\
    \hline
    Model Size (M) & 50.67 & 50.67 & 50.68 & 50.69 & 50.70 & 50.73 & 52.98 \\
    GPU Memory (MB) & 656 & 684 & 794 & 942 & 1258 & 1792 & 734 \\
    Training Time (s/step) & 0.02 & 0.02 & 0.08 & 0.08 & 0.17 & 0.18 & 0.09 \\
    Inference Time (s) & 0.17 & 0.18  & 0.27 & 0.35 & 0.44 & 0.72 & 0.22 \\
    \hline
    Success Rate (\%) & 46.1 & 45.0 & 49.1 & 49.6 & 49.6 &  40.4 & 58.3\\
    \hline
    \end{tabular}
    \end{center}
    \vspace{-0.15in}
\end{table}

\subsection{Results and Analysis}

\textbf{Main Results.} Figure \ref{fig:mikasa} shows results on two memory-intensive tasks in the MIKASA \cite{cherepanov2025memory} benchmark. Our method not only significantly outperforms its fixed-context-window counterparts, but also surpasses state-of-the-art VLA baselines by more than 20\% on average. The key to this high success rate is that the compressed episodic memory provides sufficient information to the denoising network; merely scaling the policy while conditioning on a single frame is insufficient to solve such non-Markovian tasks. Language-based (MemER) and image-based (BPP) summaries preserve enough information for the \textit{ShellGameTouch-v0} task, but remain insufficient to capture the long-range temporal dependencies required by the \textit{Long-Horizon Square} task. Figure \ref{fig:momart} shows results on the MoMaRT \cite{wong2022error} benchmark, where our proposed method achieves an average 5\% improvement over baselines.  Note that we instantiate the proposed working and episodic memory mechanism in both baselines, DP and MaIL, and it consistently improves their performance. Figure \ref{fig:momart} also demonstrates that our method performs on par with baselines on the (almost) Markovian Robomimic benchmark \cite{mandlekar2022matters}. Collectively, these results show that our proposed method effectively addresses the long-term memory challenge in non-Markovian tasks.

\textbf{Computational Efficiency.} Table \ref{tab:computation} compares the computational and storage costs of DP-PTP across different context lengths with those of DP-VPWEM on the unload-suboptimal task in the MoMaRT benchmark. While the DP-PTP model size remains nearly constant, both training and inference time increase substantially as the context length grows, whereas the success rate improves only marginally and even degrades at the context length of 128. In contrast, the additional memory module in DP-VPWEM is lightweight (around 2.24M parameters) and introduces only modest computational and storage overhead, yet achieves a higher success rate (58.3\%) than all DP-PTP variants evaluated. Besides, the VLM in BPP and MemER introduces around 0.9s latency per query, which is much more computational expensive than the lightweight compressor in VPWEM.

\textbf{Generalization.} To evaluate the generalization ability of the compressor, we transfer the frozen image encoder and compressor to unseen but related tasks. Specifically, the compressor trained on ShellGameTouch is transferred to ShellGamePush and ShellGamePick, which also require interacting with the mug hiding the ball. DP-VPWEM achieves average success rates of 90\% and 63\%, respectively, showing that the jointly trained compressor can generalize well and be directly applied to unseen tasks.

\textbf{Key Information Probe.} We perform a key information recovery experiment on the \textit{Long-Horizon Square} task to support the hippocampus analogy. We collect working and episodic memory embeddings across 100 episodes, group them by timestep indices, and train an MLP to predict the initial embedding by minimizing the mean squared error (MSE). As shown in Fig. 6(b), episodic memory preserves more initial-state information than working memory throughout the episode, although probe performance gradually decreases over time and benefits slightly from more memory tokens at later timesteps. Results suggest that the episodic memory embeddings indeed capture long-horizon information through end-to-end training, thereby improving performance on memory-intensive tasks.

\subsection{Ablation Studies}
\label{sec:exp.ablation}

\begin{figure*}[t]
    \centering
\subfloat[]{\includegraphics[width=0.73\textwidth, keepaspectratio=true,trim=20 635 20 20,clip]{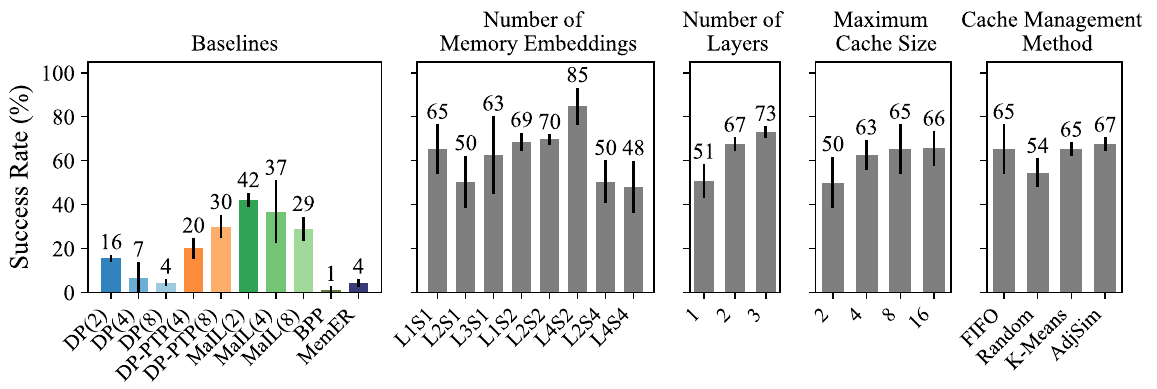}}
\subfloat[]{\includegraphics[width=0.26\textwidth, keepaspectratio=true,trim=20 635 380 10,clip]{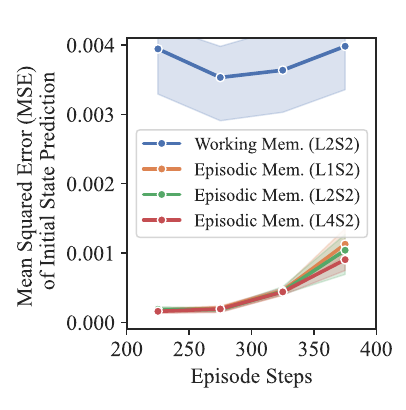}}
\caption{Ablation studies on \textit{Long-Horizon Square} task. (a) Fixed-context baselines remain limited even with longer observation horizons, while VPWEM outperforms them with only a few episodic memory embeddings. Increasing compressor depth and cache size improves performance until saturation, and cache management methods perform similarly except for random eviction. (b) Episodic memory preserves substantially more initial-state information than working memory, achieving 20$\times$ lower validation MSE. Although probe performance decreases as the episode progresses, episodic memory still yields 4$\times$ lower loss than working memory after 400 steps. More episodic memory embeddings (i.e., larger $M$) slightly improve performance at later timesteps.}
\vspace{-0.2in}
\label{fig:ablation}
\end{figure*}

We perform ablation studies on the \textit{Long-Horizon Square} task to investigate how different design choices in VPWEM affect the performance, as shown in Figure \ref{fig:ablation}(a).


\textbf{Context Length (Observation Horizon).} From the results of baselines, we observe that the performance of DP degrades as the observation horizon increases, whereas DP-PTP achieves better performance with longer contexts. MaIL outperforms DP, largely due to the recurrent nature of SSMs in the denoising network. All baselines struggle on this task because they rely on truncated observation histories. On the contrary, the L1S1 (i.e., one long-term and one short-term memory embedding) variant of our proposed method already achieves 65\% success rate, surpassing all previous baselines. Performance generally improves as the number of long-term or short-term memory embeddings increases, but may eventually degrade when too many embeddings are used. Hence, we use the L2S2 variant in our main experiments.

\textbf{Number of Layers in Compressor.} We investigate the influence of compressor depth using the L1S1 variant. The results show a clear trend: a deeper compressor improves performance by producing more informative summary embeddings and providing better conditioning for action generation. We also increase the hidden-state dimension of Mamba in MaIL from 16 to 64. However, the success rate first increases from 42\% to 47\% and then drops to 34\%, with all variants still underperforming VPWEM. This indicates that the performance gain comes not merely from adding parameters, but from the proposed memory mechanism.

\textbf{Maximum Cache Size.} We observe that performance improves as the maximum cache size increases, since more historical information is retained. However, the gains begin to saturate when the size exceeds 8. This suggests redundancy in the long observation history and indicates that the memory compressor can distill the necessary information into a small number of embeddings without sacrificing performance.

\textbf{Cache Management Method.} We also investigate several cache management methods besides FIFO: Random evicts a randomly selected embedding when the cache exceeds its maximum length; K-Means partitions embeddings into clusters by iteratively assigning embeddings to the nearest cluster center and updating the centers; and AdjSim \cite{shi2025memoryvla} computes cosine similarities between adjacent cache entries and merges the pair with the highest similarity. Results show that all methods except for Random perform similarly, hence we use the simplest FIFO in our main experiments.

\subsection{Real World Experiments}
\label{sec:exp.real}

\begin{figure}[t]
    \centering
\includegraphics[width=0.46\textwidth,keepaspectratio=true,trim=205 235 210 260,clip]{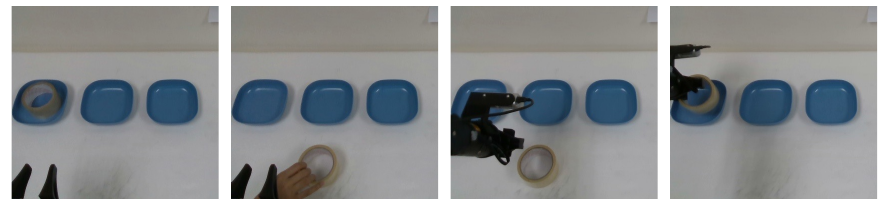}
    \vspace{-0.1in}
    \caption{Real-world robot experiments.}
    \vspace{-0.1in}
\label{fig:real_robot}
\end{figure}

To validate real-world practicality, we deploy VPWEM on an AgileX Cobot Magic platform for a memory-demanding tabletop manipulation task, as shown in Figure \ref{fig:real_robot}. The task requires the robot to return the packing tape to its original plate. We collect 180 real-robot demonstrations, train the policy for 1e6 gradient steps, and evaluate performance over 30 trials. All hyperparameters remain the same as those in Table \ref{tab:hyper}. On the real-world robot, DP-VPWEM improves the success rate from 37\% to 67\% over DP, with only a small latency increase from 72 ms to 79 ms per action chunk.

\section{Conclusion}
\label{sec:conclusion}

We presented VPWEM, a non-Markovian visuomotor policy learning framework with complementary working and episodic memories. A learnable compressor distills historical observations into compact episodic memories, enabling VPWEM to exploit long-term temporal information with bounded computational and storage costs. Extensive evaluations show substantial improvements on memory-demanding robotic tasks. Future work includes extending VPWEM to broader policy architectures, incorporating auxiliary objectives such as reconstruction, and deploying it on real-world robots across more complex tasks.





\bibliographystyle{ieeetr}
\bibliography{ref}

\end{document}